\documentclass[a4paper,fleqn]{cas-dc}

\usepackage[authoryear, round]{natbib}
\usepackage{subcaption}
\usepackage{float}
\usepackage[switch]{lineno}

\def\tsc#1{\csdef{#1}{\textsc{\lowercase{#1}}\xspace}}
\tsc{WGM}
\tsc{QE}
\tsc{EP}
\tsc{PMS}
\tsc{BEC}
\tsc{DE}
\begin{document}

% \linenumbers

\let\WriteBookmarks\relax
\def\floatpagepagefraction{1}
\def\textpagefraction{.001}

\shorttitle{Forecasting the Number of Harvest-ready Fruits of Sweet Peppers Using Multimodal Time-Series Data}

\shortauthors{E Pallotta et~al.}

\title [mode = title]{Forecasting the Number of Harvest-ready Fruits of Sweet Peppers Using Multimodal Time-Series Data}     

\author[1,2]{Enrico Pallotta}[type=editor,
                        auid=000,bioid=0,
                        prefix=,
                        orcid=0009-0003-9276-0540]
\cormark[1]


\ead{e.pallott@uni-bonn.de}

\author[1]{Mohamed Farag}[type=editor,
                        auid=000,bioid=1,
                        prefix=,
                        orcid=0000-0003-4301-1140
                        ]

\author[1]{Esra Guclu}[type=editor,
                        auid=000,bioid=2,
                        prefix=,
                        orcid=0000-0003-1106-5566
                        ]                      

\author[1,2,3]{Chris McCool}[type=editor,
                        auid=000,bioid=3,
                        prefix=,
                        orcid=0000-0002-0577-1299
                        ] 
                        
\author[1]{Ribana Roscher}[type=editor,
                        auid=000,bioid=4,
                        prefix=,
                        orcid=0000-0003-0094-6210
                        ]        

\author[1,2]{Juergen Gall}[type=editor,
                        auid=000,bioid=5,
                        prefix=,
                        orcid=0000-0002-9447-3399
                        ]

\affiliation[1]{organization={University of Bonn},
    city={Bonn},
    % citysep={}, % Uncomment if no comma needed between city and postcode
    % postcode={1043 NX}, 
    country={Germany}}

\affiliation[2]{organization={Lamarr Institute for Machine Learning and Artificial Intelligence},
    city={Bonn},
    % citysep={}, % Uncomment if no comma needed between city and postcode
    country={Germany}}

\affiliation[3]{organization={CSIRO},
    city={Pullenvale},
    % citysep={}, % Uncomment if no comma needed between city and postcode
    country={Australia}}

\begin{abstract}
Accurate yield forecasting at the individual-plant level is critical for precision agriculture and supply-chain planning, yet public datasets capturing both visual growth dynamics and per-plant measurement labels are scarce.
In this paper, we introduce a novel, annotated image time-series dataset of 691 sweet pepper plants monitored over two growing seasons, comprising 4837 images with per-plant fruit counts categorized by maturity.
We propose a multimodal deep learning framework that fuses high-dimensional image features, extracted using the DinoV3 encoder, with numerical count measurements.
Our architecture utilizes a Long Short-Term Memory (LSTM) network to model temporal dependencies and handles irregular sampling intervals common in greenhouse monitoring.
Through quantitative experiments, we demonstrate that this multimodal approach reduces RMSE over a persistence baseline by 33\% and 38\% in the 2022 and 2023 seasons, respectively, with a further 1.2\% average gain over a measurement-only model.
Furthermore, we employ Deep Ensembles and Gaussian Negative Log-Likelihood (NLL) to provide calibrated uncertainty estimates, with an Uncertainty Calibration Error (UCE) ranging from 0.39 to 0.89 depending on the cross-season evaluation direction, offering a principled confidence signal for real-world agricultural decision-making.
We release the dataset and code to support reproducible research and to accelerate development of data-driven yield forecasting methods for horticultural crops.
\end{abstract}

% Keywords

\maketitle

\section{Introduction}
Sweet pepper (\textit{Capsicum annuum}), commonly known as bell pepper, is a major horticultural crop grown worldwide for fresh-market and processing use. Global production of peppers reached a record in recent years~\cite{sood2023characterisation}, with total production measured in tens of millions of tonnes, underlining the crop's economic and food-security importance. A large and growing share of commercial sweet-pepper production takes place in controlled environments (greenhouses), where climate control, fertigation and integrated pest management enable higher yields, better fruit quality and extended production seasons compared with open-field growing. Well-managed greenhouse systems can achieve very high per-area yields and are increasingly used where year-round supply and premium quality are required.

Accurate yield forecasts in greenhouse horticulture are valuable for growers and supply-chain actors because they support operational decisions such as labour planning, harvest scheduling, storage allocation and contracts with buyers~\cite{sauviller2008predicting, lin2008neural, onoufriou2023premonition}. This need has motivated a growing body of research on forecasting methods for greenhouse crops, including mechanistic~\cite{jones1991dynamic} and data-driven~\cite{peng2023prediction} approaches and recent deep-learning methods that combine temporal models with environmental and image data~\cite{kierdorf2024investigating, kamangir2026cmavit}.

However, two limitations persist in this line of work. First, many existing studies rely on measurements and proprietary datasets collected by commercial growers that are never publicly released, limiting reproducibility and slowing the development of benchmark models for smart greenhouse management. Second, even where forecasting models are developed, they typically rely on a single modality (discrete fruit counts) and rarely provide a measure of confidence in their predictions, despite the inherently stochastic nature of individual plant development and the practical need for growers to distinguish reliable forecasts from uncertain ones.

In this work, we address both limitations. We introduce a carefully collected, multimodal dataset of sweet-pepper growth and production from a monitored greenhouse, spanning two growing seasons, 2022 and 2023, and we propose a multimodal deep learning framework for forecasting the number of harvest-ready fruits at the plant level. The framework fuses high-dimensional visual features, extracted using a DinoV3~\cite{simeoni2025dinov3} encoder, with per-plant fruit-count measurements within a Long Short-Term Memory (LSTM) network that models temporal dependencies in plant development. To accommodate the irregular sampling intervals that are common in greenhouse monitoring, the framework explicitly incorporates the time delta $\Delta_t$ between observations, allowing it to modulate temporal transitions based on elapsed time rather than assuming fixed-interval sampling. Beyond point forecasts, we extend the framework with Deep Ensembles and a Gaussian negative log-likelihood objective to produce calibrated uncertainty estimates, offering growers a principled confidence signal alongside each prediction.

Through quantitative experiments on both seasons, we show that this multimodal approach substantially outperforms a persistence baseline, cutting RMSE by 33\% in 2022 and by 38\% in 2023. It further provides a marginal but consistent gain over a counting-only model that relies solely on numerical measurements, reducing RMSE by an additional 2\% in 2022 and 0.3\% in 2023, indicating that visual features capture morphological and color transitions not fully reflected in discrete counts. We also show that the integration of temporal deltas allows the model to generalize across seasons despite irregular and varying sampling frequencies, though calibration quality is asymmetric: training on 2022 and evaluating on 2023 yields a well-calibrated ensemble (UCE of 0.39), while the reverse direction produces a higher calibration error (UCE of 0.89). Even in this less favorable direction, the resulting uncertainty estimates remain practically useful for supporting harvest-related decision-making.

By releasing both the dataset and the proposed forecasting framework, we aim to (i) enable direct comparison between forecasting methods on greenhouse sweet-pepper data, (ii) advance multimodal, image-based yield prediction by demonstrating a temporal architecture that fuses visual and numerical signals while explicitly handling irregular sampling, and (iii) provide a reproducible benchmark that couples accurate point forecasts with calibrated uncertainty quantification, supporting practical gains in labour and logistics planning for commercial growers.

\section{Related works}
\label{sec:related_works}

The intersection of computer vision and precision agriculture has shifted significantly from static trait estimation to dynamic temporal forecasting. This section reviews recent advancements in phenotypic prediction and the emerging field of generative plant growth modeling.

\subsection{Predicting phenotypes and yield forecasting}
Predicting plant traits and harvest readiness from longitudinal data is a critical component of smart greenhouse management. Early approaches often relied on manual feature extraction and statistical modeling.

\noindent For instance, \cite{das2020time} utilized neural networks to forecast morphological traits such as height, width, and areal density; however, their approach relied exclusively on geometric descriptors rather than raw image data. Similarly, \cite{buxbaum2022non} developed a dataset focused on biomass estimation across different treatment groups, highlighting the potential for time-series analysis in monitoring growth responses, though their primary focus remained on estimation rather than multi-step forecasting.

More recently, deep learning architectures have been leveraged to handle the complexity of image-based time-series. \cite{kierdorf2024investigating} introduced a method for cauliflower harvest prediction using a Multi-Layer Perceptron (MLP) trained on features extracted from a ResNet backbone, integrated with positional embeddings to encode temporal information. While effective for single-stage readiness, this approach does not explicitly model the sequential dependencies between growth stages. 

In the domain of yield prediction, \cite{de2025strawberry} proposed an LSTM-based framework to iteratively predict the count of harvest-ready strawberries. A key contribution of their work is the integration of physical constraints into the loss function, ensuring that the transition between ripeness classes adheres to biological growth rates. However, their model relies solely on count-based inputs, discarding potentially rich visual information present in the raw images. 

Most relevant to our study is the work of \cite{shimomoto2025predicting}, who addressed sweet pepper yield forecasting. They employed a two-stage pipeline consisting of fruit detection and color-based classification, followed by a linear regression model to predict the next week's yield. While their setup provides a foundational baseline for pepper forecasting, our work extends this by exploring non-linear deep learning architectures and investigating the benefits of multimodal fusion, where both discrete counts and visual features are utilized for prediction.

\subsection{Generative modeling for plant growth}
A parallel line of research focuses on ``predicting the future'' through image synthesis, where the goal is to generate the visual state of a plant at a future time step. \cite{yasrab2021predicting} utilized Generative Adversarial Networks (GANs) to predict future frames of \textit{Arabidopsis thaliana} root growth from segmentation masks. While promising, GAN-based approaches in agriculture have often been critiqued for producing blurry results that lack the high-frequency detail required for precise phenotypic analysis \cite{wang2022predicting, drees2021temporal}.

To address these limitations, \cite{wang2022predicting} proposed the use of Spatio-Temporal LSTMs (ST-LSTM) combined with Memory In Motion (MIM) modules to predict wheat growth sequences. By framing the problem as a video prediction task, they achieved higher structural similarity (SSIM) compared to standard convolutional LSTMs. However, these generative models primarily focus on visual appearance rather than the explicit quantification of harvestable yield. Our work bridges this gap by utilizing temporal visual data not for image synthesis, but as a high-dimensional feature space to improve the accuracy of numerical yield forecasting.

\subsection{Statistical and model-based approaches}
Traditional agronomic models continue to play a role in phenotype prediction. \cite{strawberrymodel} developed a parametrized model based on the correlation between environmental variables (e.g., temperature, humidity) and physical traits like plant height. While these models offer high interpretability, they often struggle to capture the stochastic nature of individual plant development compared to data-driven computer vision approaches. Our framework aims to combine the strengths of data-driven learning with the temporal structure inherent in biological growth.

\subsection{Uncertainty quantification} 

Due to its significance, there have been multiple attempts to integrate uncertainty quantification techniques into the agricultural machine learning framework. \cite{wang2019pattern} use a full Bayesian approach to estimate uncertainty for subsoil heterogeneity estimations. They specify a prior distribution and use Bayes' rule to get the posterior as they use a model with a relatively low number of parameters compared to deep learning models. MC-Dropout and Stochastic Gradient Langevin Dynamics (SGLD) \cite{Welling2011BayesianLV} have been used by \cite{HERNANDEZ2020106597} as an approximation of a full Bayesian approach for plant disease detection. \cite{modelling2040040} and \cite{meenken2021bayesian} have investigated Bayesian approaches for real-time crop management and estimating crop sensitivity to weather changes, using hybrid modeling that combines machine learning with mechanistic models. 

Several works \cite{modelling2040040, PADARIAN2022116063, LI2023108414, Celikkan_2023_ICCV} have focused on using MC-Dropout or ensembles to estimate data uncertainty or model uncertainty in deep learning models for applications in agricultural or soil modeling. These approaches are computationally efficient during training and provide a direct way to measure uncertainty estimates by using metrics such as mean and variance. 
As an alternative, conformal prediction has gained attention for agricultural applications \cite{melki2023group, Farag2023} due to its simplicity and robust statistical guarantees for providing calibrated uncertainty estimates. 

\section{Material and methods}
\subsection{Data acquisition}
\begin{figure}
        \centering
        \includegraphics[width=1\linewidth]{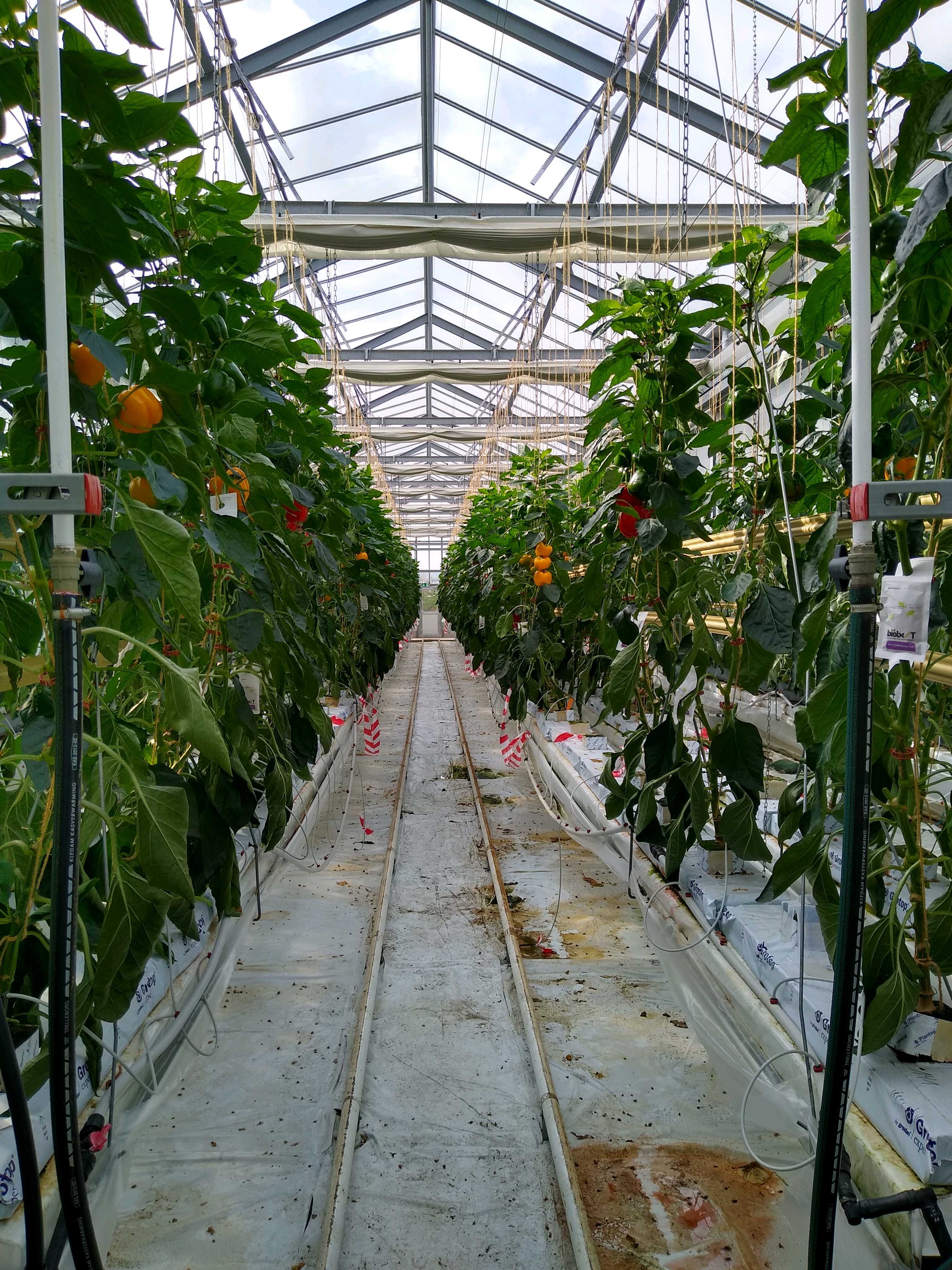}
        \caption{Glasshouse at the University of Bonn Campus Klein Altendorf.}
        \label{fig:glasshouse}
\end{figure}

\begin{figure}
        \centering
        \includegraphics[width=1\linewidth]{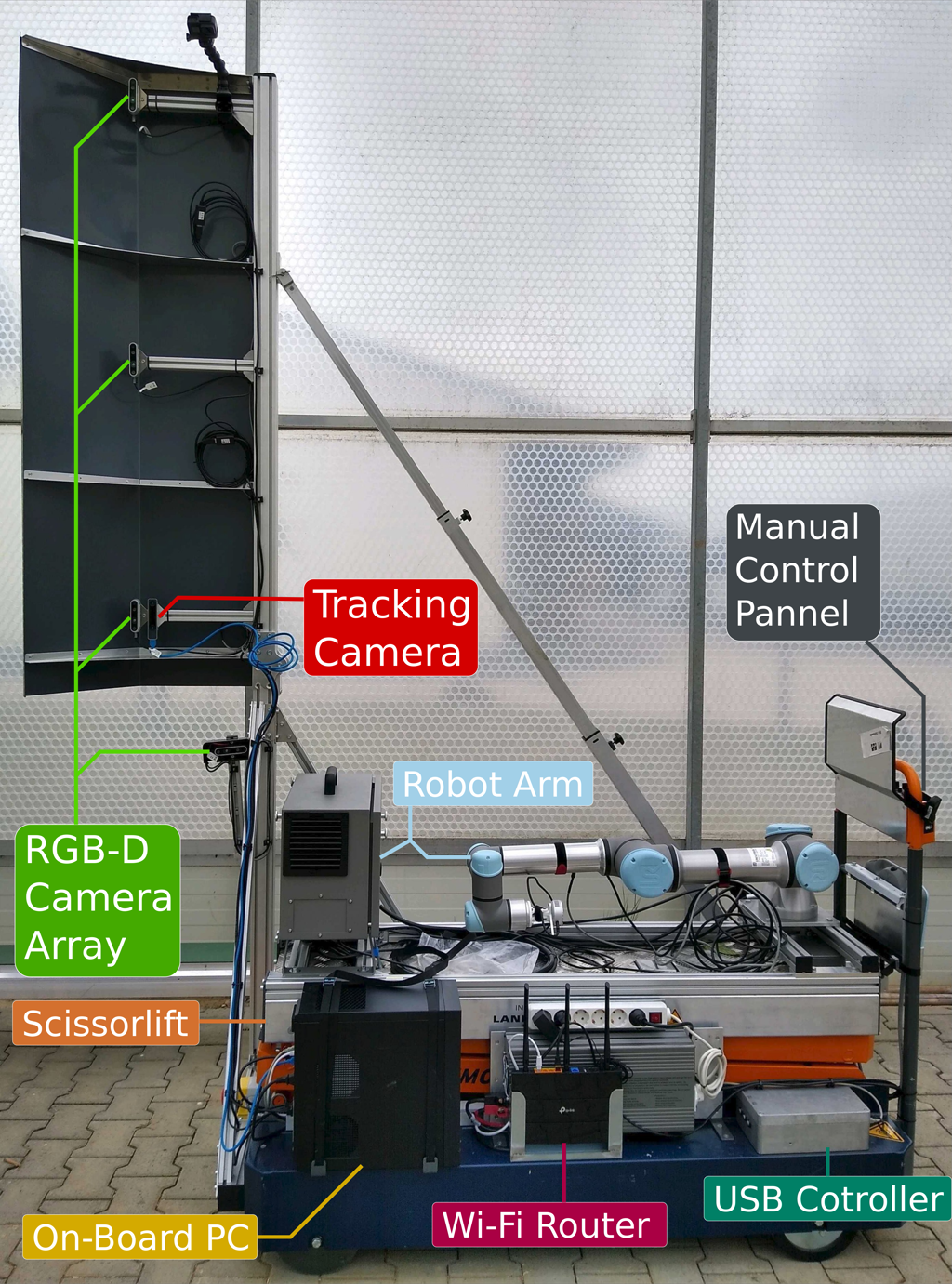}
        \caption{The PATHoBot robotic platform used for capturing image data.}
        \label{fig:pathobot}
\end{figure}

The data utilized in this study were captured during the 2022 and 2023 seasons within the sweet pepper chamber of a commercial glasshouse (Figure~\ref{fig:glasshouse}) at the University of Bonn Campus Klein Altendorf (CKA). 
The chamber consists of 6 rows, each approximately 34 meters in length, where \textit{mavera} and \textit{allrounder} sweet peppers are grown. 
Data acquisition was performed using the robotic platform PATHoBot \cite{smitt2021pathobot} shown in Figure~\ref{fig:pathobot}, which is designed specifically for monitoring and intervention tasks in glasshouse environments. 
PATHoBot operates by moving along the pipe-rail system in the glasshouse and is equipped with three Intel RealSense D435i RGB-D cameras, enabling the capture of synchronized color and depth imagery. 
In addition to visual data, the platform provides wheel odometry information during operation. 
For each of the two seasons, data were captured on 7 days, resulting in a total of 14 recording days across both years.

\subsection{Time series annotation}
As a result of the data acquisition process, we have one video for each row and recording date, which  is stored as a sequence of images. In order to create a per-plant image timeseries dataset, we aligned the recorded images over the growing period. To do so, we designed a python GUI that allowed us to go through a pair of sequences and save aligned image pairs manually. Images were selected such that they would contain most of the plant's visible parts, furthermore an overlayed grid allowed to achieve a precise spatial alignment between plant frames over different dates. After repeating this process for both 2022 and 2023 recordings, we obtained visual timeseries data for 691 plants, each timeseries representing the growth over 7 dates, resulting in 4837 image frames.

\subsection{Plant-level fruits annotation}\label{fruits_annotation}
To generate instance-level sweet pepper annotations for selected image time series, we employ CTVIS~\cite{ying2023ctvis}, a video-based instance segmentation model.
Instead of relying on image-based models operating on individual frames, a video instance segmentation (VIS) framework is chosen to leverage the richer spatial-temporal information available in sequential data.
By processing temporally adjacent frames jointly, VIS models can exploit multiple viewpoints of the same object over time, resulting in more robust feature representations and more reliable instance association.

Although the target data for annotation consists of 4837 individual reference frames extracted from the 2022 and 2023 recordings, the original data is inherently sequential.
To enable the application of the video-based model, the original recordings are divided into subsequences of 30 consecutive frames.
Each subsequence has a selected reference frame at its center.
In this way, a total of 4837 short video clips are formed, each centered around a reference frame, allowing the VIS model to exploit temporal context while producing annotations for the target images.

Spatial-CTVIS is trained on BUP-ST20~\cite{guclu2025weakly}, a weakly labelled spatial-temporal dataset of sweet peppers suitable for robotic perception in agricultural scenarios.
For each reference frame, the model generates instance segmentation masks, class labels (red, green, yellow, mixed-red, and mixed-yellow), and associated confidence scores for each sweet pepper in the image.
These outputs serve as the final annotations for the image time series utilized in this study.

\begin{figure}
    \centering
    \begin{subfigure}{\linewidth}
        \centering
        \includegraphics[width=\linewidth]{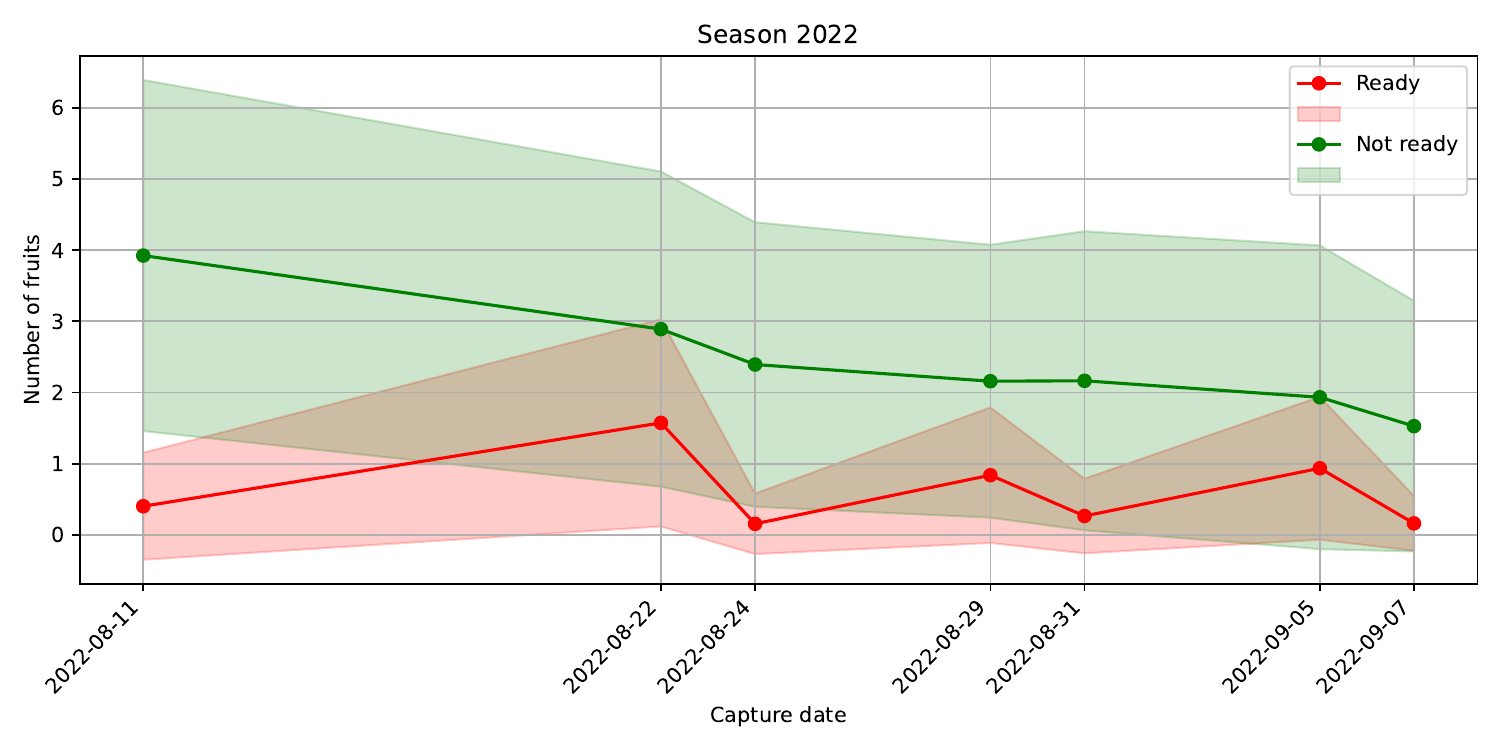}
        \caption{}
        \label{fig:timeseries_2022}
    \end{subfigure}
    \hfill
    \begin{subfigure}{\linewidth}
        \centering
        \includegraphics[width=\linewidth]{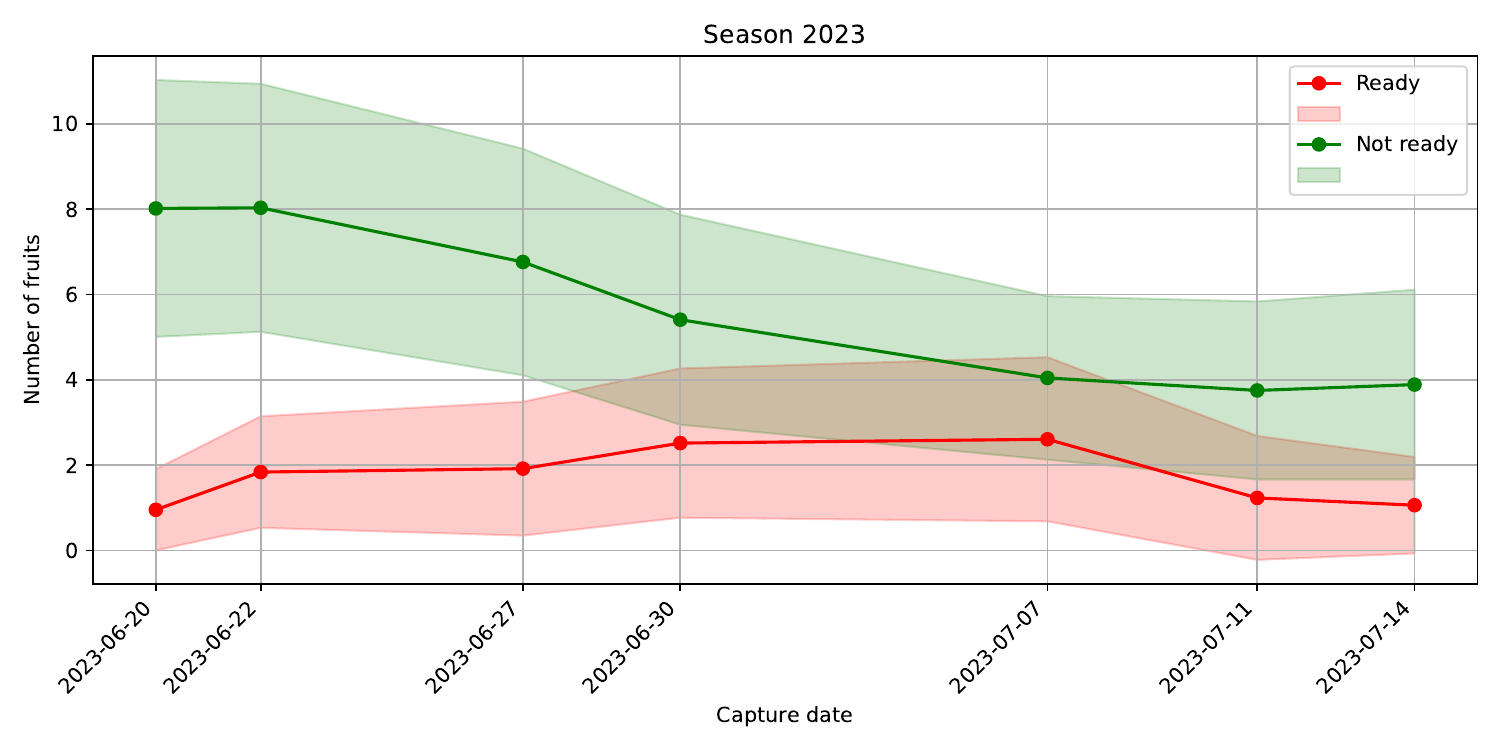}
        \caption{}
        \label{fig:timeseries_2023}
    \end{subfigure}
    \caption{Timeseries showing the number of ready and not ready fruits for years 2022 and 2023.}
    \label{fig:timeseries_stats}
\end{figure}

As illustrated in Figure \ref{fig:timeseries_stats}, the timeseries obtained for 2022 and 2023, where the various classes have been grouped into “ready” and “not ready”, display comparable trends, even though the measurement frequency differs considerably between the two years.
Specifically, the green and mixed categories (not ready) gradually decline over time as the fruits develop, whereas the ready category fluctuates due to human intervention, since fruits are harvested when needed.

\subsection{Forecasting framework}
\begin{figure*}
    \centering
    \includegraphics[width=\linewidth]{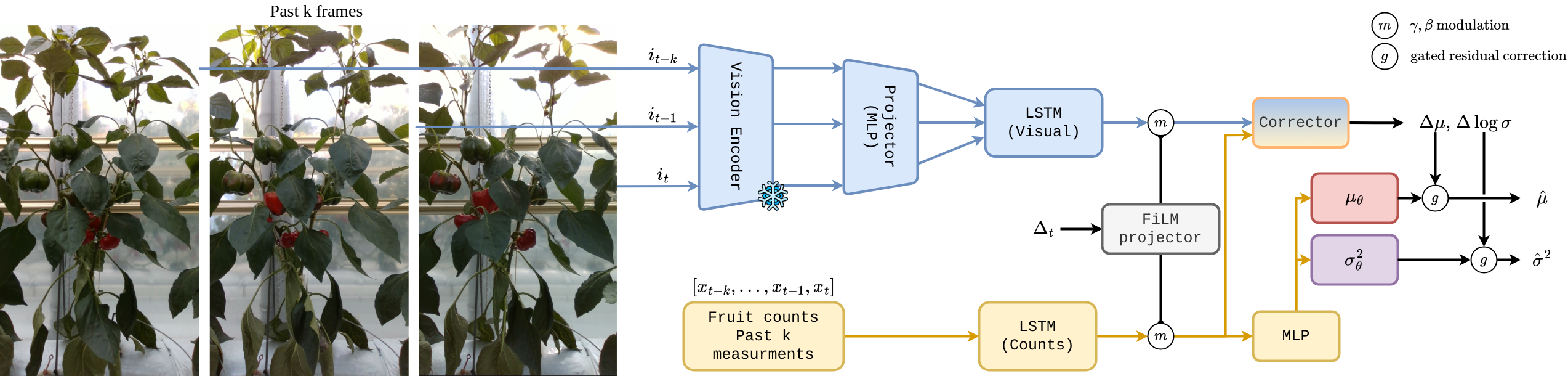}
    \caption{Architecture of the proposed model. Past frames $i_{t-k},\dots,i_t$ are encoded by a frozen vision encoder, projected by an MLP, and modeled by a visual LSTM branch, while the corresponding fruit-count measurements $[x_{t-k},\dots,x_t]$ are processed by a parallel counting LSTM branch. The forecast horizon $\Delta_t$ is mapped by a FiLM projector to per-channel scale-and-shift parameters that modulate ($m$) both LSTM context embeddings. The modulated counting context is passed through an MLP to produce a base prediction $(\mu_\theta,\sigma_\theta^2)$, while both modulated contexts are jointly fed to a corrector that estimates a residual correction $(\Delta\mu,\Delta\log\sigma)$. A learned gate ($g$) combines the base prediction with this correction to yield the final predictive mean $\hat{\mu}$ and variance $\hat{\sigma}^2$ that will be used in the ensemble to estimate $\hat{y}_{t+1}$.}
    \label{fig:main_fig}
\end{figure*}

To transform raw counting information into actionable agricultural insights, we formulated a forecasting task focused on harvest readiness. We aggregated the five maturity classes into two primary categories:
\begin{enumerate}
    \item \textbf{Ready to Harvest:} Comprising the \textit{red} and \textit{yellow} classes.
    \item \textbf{Not Ready:} Comprising the \textit{green}, \textit{mixed-red}, and \textit{mixed-yellow} classes.
\end{enumerate}

The objective is to predict the number of fruits ready for harvest at a future time step $t+1$, given the historical state of the plant. We define the forecasting function $f_{\theta}$ as:
\begin{equation}
    f_{\theta}(x_t, x_{t-1}, \dots, x_{t-k}, \Delta_t) = \hat{y}_{t+1}\,,
\end{equation}
where $x_t$ represents the state of the plant at time $t$ (e.g., fruit counts per class), $k$ is the number of past observations, $\Delta_t$ represents the number of days between the most recent observation $t$ and the target date $t+1$ and $\hat{y}_{t+1}$ is the predicted count of harvest-ready fruits at the target date.

This forecasting problem naturally takes the form of a multivariate time-series prediction task, in which plant development follows a gradual and temporally dependent process. Fruit appearance, growth, and ripening are driven by cumulative effects over time, such as past fruit load and growth dynamics, which cannot be captured by models that assume independent observations. For this reason, we employ Long Short-Term Memory (LSTM)~\cite{graves2012long} modules to build our network $f_{\theta}$. LSTMs are a class of recurrent neural networks specifically designed to capture long-term temporal dependencies through gated memory mechanisms. This makes them well suited for agricultural growth modeling, where the current yield state depends not only on recent observations but also on longer-term developmental trends. Moreover, LSTMs can handle variable-length sequences and irregular temporal sampling when augmented with explicit temporal information, such as the time delta $\Delta_t$. This is particularly relevant in greenhouse monitoring scenarios, where image acquisition and annotations may not occur at fixed intervals. By explicitly providing $\Delta_t$ as an input, the model can learn to modulate temporal transitions based on the elapsed time between observations.

Rather than injecting the elapsed time $\Delta_t$ as a raw, concatenated input, we condition the model's internal temporal representations on $\Delta_t$ through a Feature-wise Linear Modulation (FiLM)~\cite{perez2018film} mechanism, described in Section~\ref{sec:mm_iputs}. This design allows the network to learn \emph{how} to reshape its representations as a function of elapsed time, rather than treating $\Delta_t$ as an undifferentiated additional input dimension.

The model parameters $\theta$ are optimized using gradient descent to minimize an objective function that quantifies the discrepancy between the predicted and ground-truth number of harvest-ready fruits. Given a training set of $N$ samples, the parameters are iteratively updated to minimize the total loss:
\begin{equation}\label{eq:detloss}
    \mathcal{L}(\theta) = \frac{1}{N} \sum_{i=1}^{N} \ell \left( y^{(i)}, \hat{y}^{(i)} \right)\,,
\end{equation}
where $\ell$ represents the specific loss criterion employed during training, the details of which are provided in Section \ref{sec:loss}.

\subsubsection{Multimodal inputs}\label{sec:mm_iputs}
While fruit counts are important, they do not reflect the full state of the plant. We therefore investigate the predictive value of visual information by augmenting a counting-only backbone with a parallel branch operating on plant images $i_t$. Our architecture, illustrated in Figure~\ref{fig:main_fig}, processes them in two separate temporal branches that are combined at the prediction stage.

The \textbf{counting branch} feeds the sequence of past fruit counts $[x_{t-k}, \dots, x_t]$ directly to a dedicated LSTM, producing a temporal context $h_m$ that, on its own, corresponds to a purely count-based forecaster.

The \textbf{visual branch} extracts deep visual features from each past frame $i_j$, $j \in \{t-k,\dots,t\}$, using the frozen DinoV3 foundation model \cite{simeoni2025dinov3}. DinoV3 is a Vision Transformer \cite{dosovitskiy2020image} trained through self-supervised learning, which allows it to learn complex visual patterns and object shapes directly from vast datasets without requiring manual human labels. By comparing different ``views'' of the same image, the model develops an intuitive understanding of the underlying structure and consistency of objects, enabling the extraction of robust and semantically meaningful representations. The resulting features are $\ell_2$-normalized, projected to a lower-dimensional space by a small MLP, and passed to a second, dedicated LSTM, yielding a visual temporal context $h_v$.

Since the elapsed time $\Delta_t$ between the last observation and the forecast date can vary, both contexts are modulated by $\Delta_t$ prior to fusion: a lightweight FiLM conditioner maps $\Delta_t$ to per-channel scale and shift parameters $(\gamma, \beta)$, which rescale each context as $h' = h \odot (1+\gamma) + \beta$. The conditioner's output layer is zero-initialized, so at initialization $\gamma = \beta = 0$ and both contexts pass through unmodulated; because this starting point is the identity mapping, the model is not forced to condition on $\Delta_t$.

The modulated counting context $h'_m$ is then passed through a small MLP head that
produces a base prediction of the target's mean and log-variance:
\begin{equation}
\begin{split}
z_m &= \mathrm{ReLU}\big(W_b h'_m + b_b\big), \\
\mu_{\theta} &= W_\mu z_m + b_\mu, \\
\log\sigma^2_{\theta} &= W_\sigma z_m + b_\sigma.
\end{split}
\label{eq:base_head}
\end{equation}

In parallel, $h'_m$ and the modulated visual context $h'_v$ are jointly passed
to a corrector network -- an MLP applied to their concatenation
$[h'_m; h'_v]$ -- that predicts a residual adjustment to this base prediction:
\begin{equation}
\begin{split}
z_c &= \mathrm{ReLU}\big(W_{c}[h'_m; h'_v] + b_{c}\big), \\
\Delta \mu &= W_{c,\mu} z_c + b_{c,\mu}, \\
\Delta \log\sigma^2 &= W_{c,\sigma} z_c + b_{c,\sigma}.
%(\Delta\mu,\ \Delta\log\sigma^2) &= W_{c,\sigma} z_c + b_{c,\sigma}.
\end{split}
\label{eq:corrector}
\end{equation}

A learned scalar gate $g=\tanh(\alpha)$, also zero-initialized so that $g=0$
at the start of training, weights the contribution of this residual, giving
the model's final predictive mean and log-variance:
\begin{equation}
\begin{split}
\hat\mu &= \mu_{\theta} + g\,\Delta\mu, \\
\log\hat\sigma^2 &= \log\sigma^2_{\theta}
+ g\,\Delta\log\sigma^2.
\end{split}
\label{eq:gated_output}
\end{equation}
This gated-residual design guarantees that, at initialization ($g=0$), the
model is exactly equivalent to a counting-only forecaster, and that visual
information is incorporated only to the extent that it improves predictive
accuracy.
The specific output quantities (predictive mean and variance) produced by this fusion are detailed in Section~3.4.3.

\subsubsection{Ensemble learning for uncertainty quantification}

Reliable fruit count predictions alone are insufficient for practical deployment; a model that also quantifies its own uncertainty enables growers to make more informed harvesting decisions, particularly under ambiguous or out-of-distribution conditions. To this end, we extend the deterministic forecasting framework described above to produce both a point estimate and a calibrated uncertainty measure for each prediction.

We adopt Deep Ensembles~\cite{Lakshminarayanan2017} as a tractable approximation to Bayesian Neural Networks (BNNs)~\cite{Abdar2021}. Rather than maintaining a full posterior distribution over model parameters, which is computationally prohibitive for sequence models, Deep Ensembles achieve uncertainty estimation by training $M$ independent copies (members) of $f_{\theta}$, each initialized with a distinct random seed. The diversity induced by different initializations encourages the members to explore different regions of the loss landscape, yielding a set of predictions whose disagreement reflects the model's uncertainty. The ensemble prediction is then obtained by averaging across all members:

\begin{equation}
    \tilde{f}_{\theta}(x_t, \dots, x_{t-k}, \Delta_t) = \frac{1}{M} \sum_{m=1}^{M} f_{\theta}^{(m)}(x_t, \dots, x_{t-k}, \Delta_t)\,,
\end{equation}

where the predictive variance across members serves as a proxy for model uncertainty, providing a principled signal for downstream decision-making in the greenhouse monitoring pipeline.

\subsubsection{ Probabilistic forecasting with Gaussian negative log-likelihood} \label{sec:loss}

While the MSE loss~\eqref{eq:detloss} yields deterministic point estimates of $\hat{y}_{t+1}$, it is not designed to capture predictive uncertainty by using the variance of the ensemble mean prediction. To address this, we adopt the probabilistic formulation of~\cite{Lakshminarayanan2017}, which extends the network to model a full predictive distribution rather than a single output value.

Specifically, each ensemble member $f_{\theta}^{(m)}$ is modified to output two quantities: a predicted mean $\mu_{\theta}^{(m)}(x_t, \dots, x_{t-k}, \Delta_t)$ and a predicted variance $\sigma^{2(m)}_{\theta}(x_t, \dots, x_{t-k}, \Delta_t) > 0$, parameterizing a Gaussian distribution over the target fruit count. The model is then trained by minimizing the Gaussian Negative Log-Likelihood (NLL), which simultaneously penalizes inaccurate mean predictions and learns prediction variance:
\begin{equation}
    \mathcal{L}_{\text{NLL}}^{(m)} = \frac{1}{N} \sum_{i=1}^{N} \left[ \frac{\left( y_{t+1}^{(i)} - \mu_{\theta}^{(m)(i)} \right)^2}{2\sigma^{2(m)(i)}_{\theta}} + \frac{1}{2} \log \sigma^{2(m)(i)}_{\theta} \right]\,.
\end{equation}
Crucially, this loss prevents the model from trivially reducing uncertainty by inflating $\sigma^2$: any increase in predicted variance is penalized by the $\log \sigma^2$ term, while underestimating variance is penalized by the quadratic residual term. To ensure numerical stability and enforce positivity, we pass the raw variance output through a softplus activation and apply clamping.

The final ensemble prediction is then obtained as a mixture of $M=4$ Gaussian distributions, one per member, approximated by a single Gaussian whose mean and variance are:
\begin{equation}
    \hat{\mu} = \frac{1}{M} \sum_{m=1}^{M} \mu_{\theta}^{(m)}, \quad
    \hat{\sigma}^{2} = \frac{1}{M} \sum_{m=1}^{M} \left( \sigma^{2(m)}_{\theta} + {\mu_{\theta}^{(m)}}^2 \right) - {\hat{\mu}}^2
\end{equation}
where $\hat{\mu}$ serves as the final fruit count estimate and $\hat{\sigma}^{2}$ captures both the aleatoric uncertainty (averaged from each member's learned variance) and the epistemic uncertainty (from disagreement between member means).

\subsection{Experimental protocol}
To evaluate the model’s ability to generalize across different growing seasons, we adopt a leave-one-year-out cross-validation strategy. Specifically, the model is trained and validated on data from one year using a 95/5\% split and evaluated on the held-out year. This procedure is performed twice, using 2022 for training and validation and 2023 for testing, and vice versa. This evaluation protocol ensures that forecasting performance is robust to inter-annual variations in environmental conditions, greenhouse management practices, and plant growth dynamics.

We train two variants of the LSTM model:
\begin{itemize}
    \item \textit{Counting}: a standard model that uses only historical fruit-count information of each plant as input.
    \item \textit{Multimodal}: a model that combines fruit-count data and extracted image features.
\end{itemize}
If not otherwise specified, we use $k=5$ past measurements as observations. 

We report performance using two complementary point-forecast metrics. The Root Mean Squared Error (RMSE) is defined as:

\begin{equation}
    \mathrm{RMSE} = \sqrt{\frac{1}{N}\sum_{i=1}^{N}\left(y^{(i)}_{t+1} - \hat{y}^{(i)}_{t+1}\right)^2},
    \label{eq:rmse}
\end{equation}

where $N$ is the number of forecast instances in the test set. RMSE is
expressed in the same units as the fruit counts and penalizes large
deviations disproportionately, making it sensitive to occasional severe
mis-counts.

To assess accuracy relative to a naive reference and to allow comparison
across seasons with different baseline difficulty, we additionally report the Mean Absolute Scaled Error (MASE)~\cite{hyndman2006another}:

\begin{equation}
    \mathrm{MASE} = \frac{\dfrac{1}{N}\sum_{i=1}^{N}\left|y^{(i)}_{t+1}-\hat{y}^{(i)}_{t+1}\right|}
    {\dfrac{1}{N_{\text{train}}-1}\sum_{j=2}^{N_{\text{train}}}\left|y^{(j)}_{\text{train}} - y^{(j-1)}_{\text{train}}\right|},
    \label{eq:mase}
\end{equation}
where the numerator is the mean absolute forecast error on the test set,
and the denominator is the mean absolute one-step difference of the
training-year series (written as a single long timeseries for a simpler equation), i.e. the in-sample error of a naive persistence
forecaster.
A MASE below one indicates the model improves on the naive persistence forecaster in absolute-error terms, while values above one indicate the opposite.
Being scale-independent, MASE also enables direct comparison of forecast
quality between the 2022 and 2023 seasons despite their differing count
magnitudes and baseline RMSE.

To isolate the contribution of the elapsed-time signal $\Delta_t$ within the Multimodal model, we additionally train three variants that differ only in how $\Delta_t$ is incorporated: omitting it entirely, concatenating it as a raw scalar input to the fused representation, and the proposed FiLM-based conditioning. This ablation is reported and discussed in Section~\ref{sec:delta-ablation}.

Finally, we evaluate the effect of temporal context by varying the number of past available states. Given that each time series consists of seven observation dates, we train model variants using between one and six past states as input.

\subsection{Technical details}
The extraction of the image features is done using the DinoV3-ViT-H+ vision encoder.
For training, we used AdamW optimizer with a starting learning rate of $10^{-3}$ to $10^{-4}$ (optimized via hyper-parameter search on the validation set) and weight decay of $10^{-5}$. A reduce-on-plateau mechanism was added to the learning rate scheduler and early stopping was used to prevent overfitting.
All the experiments were run on a single NVIDIA A40 (48 GB RAM) with the following libraries setup: CUDA (12.6), Python (3.10.12), PyTorch (2.7.1).

\begin{table}[t]
    \centering
    \resizebox{\linewidth}{!}{%
    \begin{tabular}{l|cc|cc}
         & \multicolumn{2}{c}{2022} & \multicolumn{2}{c}{2023} \\
         Method & RMSE & MASE & RMSE & MASE \\
         \midrule
         Baseline & 1.261 & -- & 2.066 & --\\
         Counting & \underline{0.866} & \underline{0.628} & \underline{1.287} & \underline{0.785} \\
         Multimodal & \textbf{0.848} & \textbf{0.622} & \textbf{1.283} & \textbf{0.770} \\
    \end{tabular}
    }
    \caption{Comparison of our multimodal forecasting model against a model using just counting values and a persistency prediction baseline.}
    \label{tab:main_results}
\end{table}

\begin{table}[t]
    \centering
    \resizebox{\linewidth}{!}{%
    \begin{tabular}{l|cc|cc}
          & \multicolumn{2}{c}{2022} & \multicolumn{2}{c}{2023} \\
         $k=5$ & RMSE & MASE & RMSE & MASE \\
         \midrule
         Baseline & 1.261 & -- & 2.066 & --\\
         \textit{Without} $\Delta_t$ & 0.894 & 0.626 & \textbf{1.283} & \textbf{0.738} \\
         $\Delta_t$ \textit{concat} & \underline{0.865} & \textbf{0.619} & 1.296 & \underline{0.746} \\
         $\Delta_t$ FiLM & \textbf{0.848} & \underline{0.622} & \textbf{1.283} & 0.770 \\
    \end{tabular}
    }
    \caption{Ablation of the temporal delta ($\Delta_t$) encoding strategy within the multimodal forecasting model. \emph{Without $\Delta_t$}, the elapsed-time signal is omitted entirely; \emph{$\Delta_t$ concat} appends $\Delta_t$ as a raw scalar input to the fused representation; \emph{$\Delta_t$ FiLM} uses the proposed FiLM-based conditioning (Section~3.4.1). The $\Delta_t$ FiLM row corresponds to the Multimodal model reported in Table~\ref{tab:main_results}.}
    \label{tab:delta-ablation}
\end{table}

\section{Results and discussion}
\subsection{Quantitative performance analysis}
The experimental results, summarized in Table 1, demonstrate that our deep learning framework significantly outperforms the persistence baseline across both growing seasons. In the \textit{season 2022}, the Multimodal model achieved the highest accuracy, with an RMSE of 0.848 and a MASE of 0.622, corresponding to a 32.7\% RMSE reduction over the baseline (1.261). In the \textit{season 2023}, both models yielded a substantially larger relative improvement over the season's baseline (2.066), reducing RMSE by approximately 38\%. In both years, the inclusion of DinoV3-extracted visual features provided a consistent gain over the Counting-only model. In contrast to counts, the high-dimensional image features can capture subtle morphological and color transitions, such as the gradual ripening of a "mixed" pepper.

\subsection{Impact of temporal delta encoding}\label{sec:delta-ablation}
To assess how the elapsed time $\Delta_t$ between the last observation and the forecast date should be incorporated into the multimodal model, we compare three variants in Table~\ref{tab:delta-ablation}: omitting $\Delta_t$ entirely (\emph{Without $\Delta_t$}), concatenating $\Delta_t$ as a raw scalar input to the fused representation (\emph{$\Delta_t$ concat}), and the proposed FiLM-based conditioning described in Section~3.4.1 (\emph{$\Delta_t$ FiLM}).

Explicit $\Delta_t$ conditioning is not uniformly beneficial: FiLM yields the best RMSE in 2022 and ties for best in 2023, but its MASE is the worst of the three variants in both seasons, with the no-$\Delta_t$ variant instead achieving the lowest MASE in 2023. Concatenation sits between the two on most metrics. We adopt FiLM as our default given its consistent RMSE advantage, while noting this trade-off rather than presenting FiLM as a uniform improvement.

\begin{figure*}[t]
    \centering
    \begin{subfigure}{.48\linewidth}
        \centering
        \includegraphics[width=\linewidth]{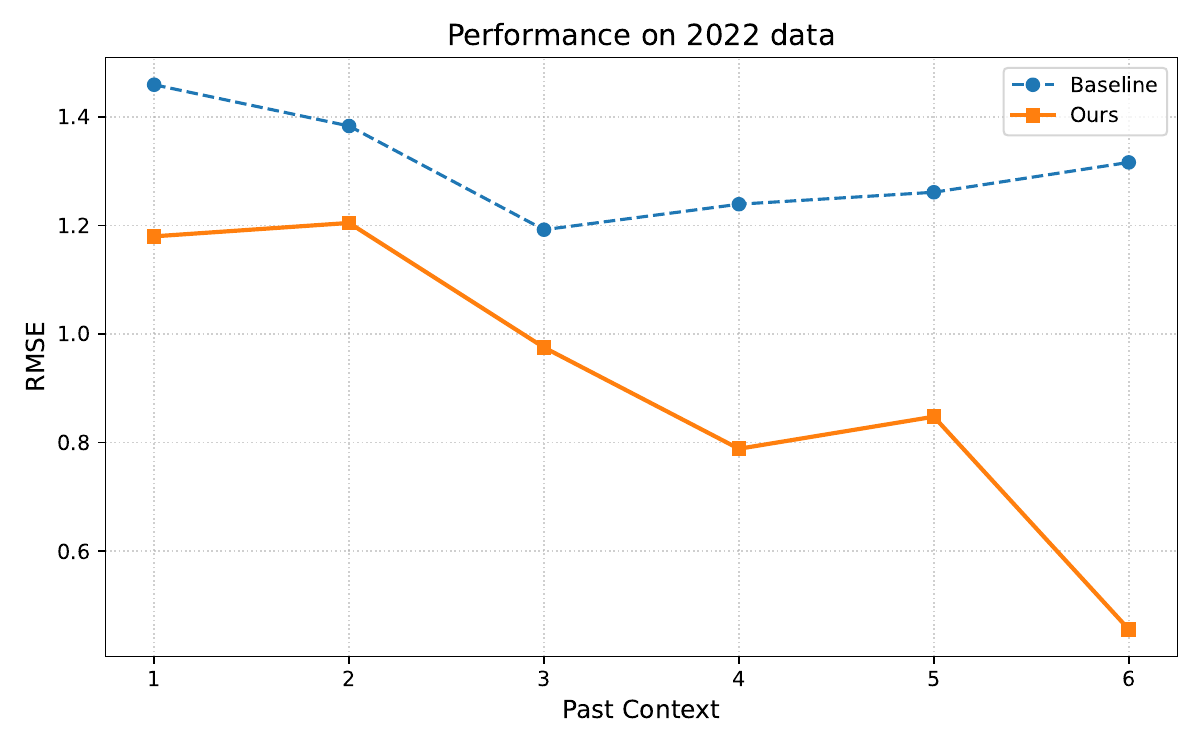}
        \caption{}
        \label{fig:2022_test_res}
    \end{subfigure}
    \hfill
    \begin{subfigure}{.48\linewidth}
        \centering
        \includegraphics[width=\linewidth]{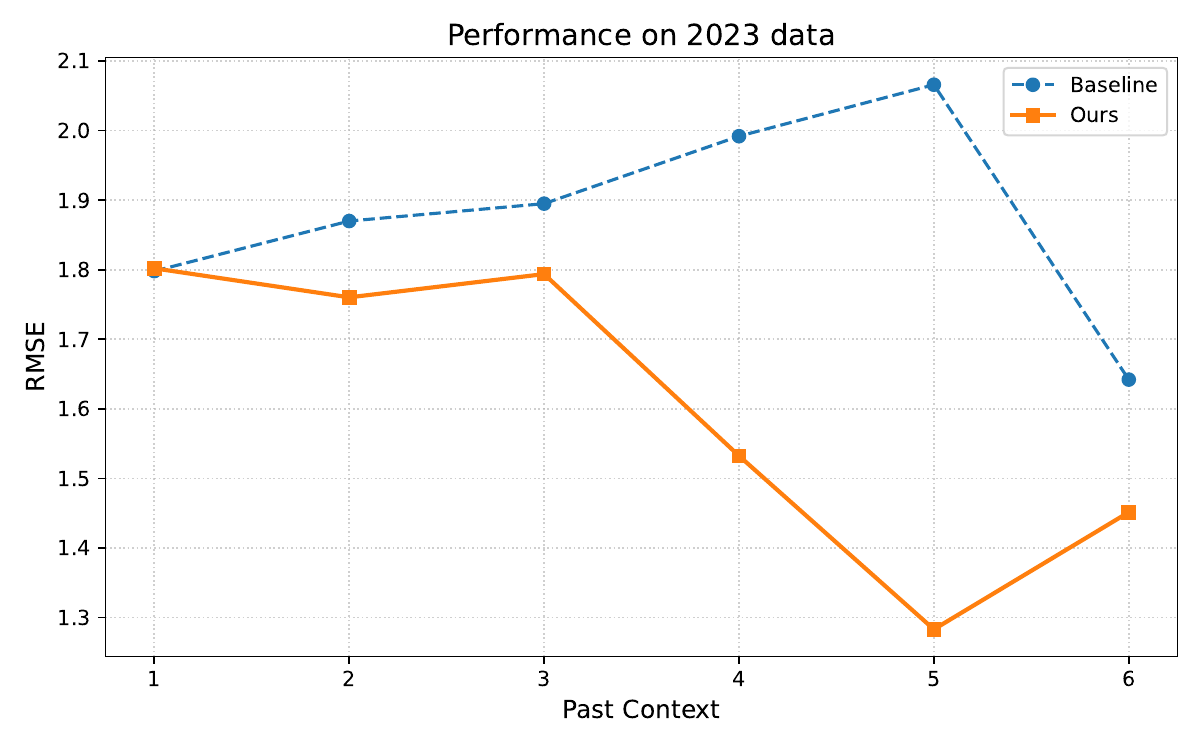}
        \caption{}
        \label{fig:2023_test_res}
    \end{subfigure}
    \caption{RMSE performance varying the number $k$ of past observations in comparison to the persistence baseline.}
    \label{fig:test_res_over_time}
\end{figure*}

\subsection{Impact of temporal context}
The influence of the number ($k$) of past observations on forecasting performances is reported in Figure 5. 
The general trend is clear, increasing the number of past observations, i.e., past context information, leads to better performance. We can see a nearly linear reduction in RMSE for the evaluation on the 2022 season. On the other hand, for 2023, a distinct performance pivot occurs after three days of context.
Beyond this point, the RMSE for our model drops sharply, suggesting that the LSTM requires a minimum historical sequence to accurately model the non-linear growth dynamics of sweet peppers under irregular sampling conditions.

\begin{figure*}[t]
    \centering
    \begin{subfigure}{.48\linewidth}
        \centering
        \includegraphics[width=\linewidth]{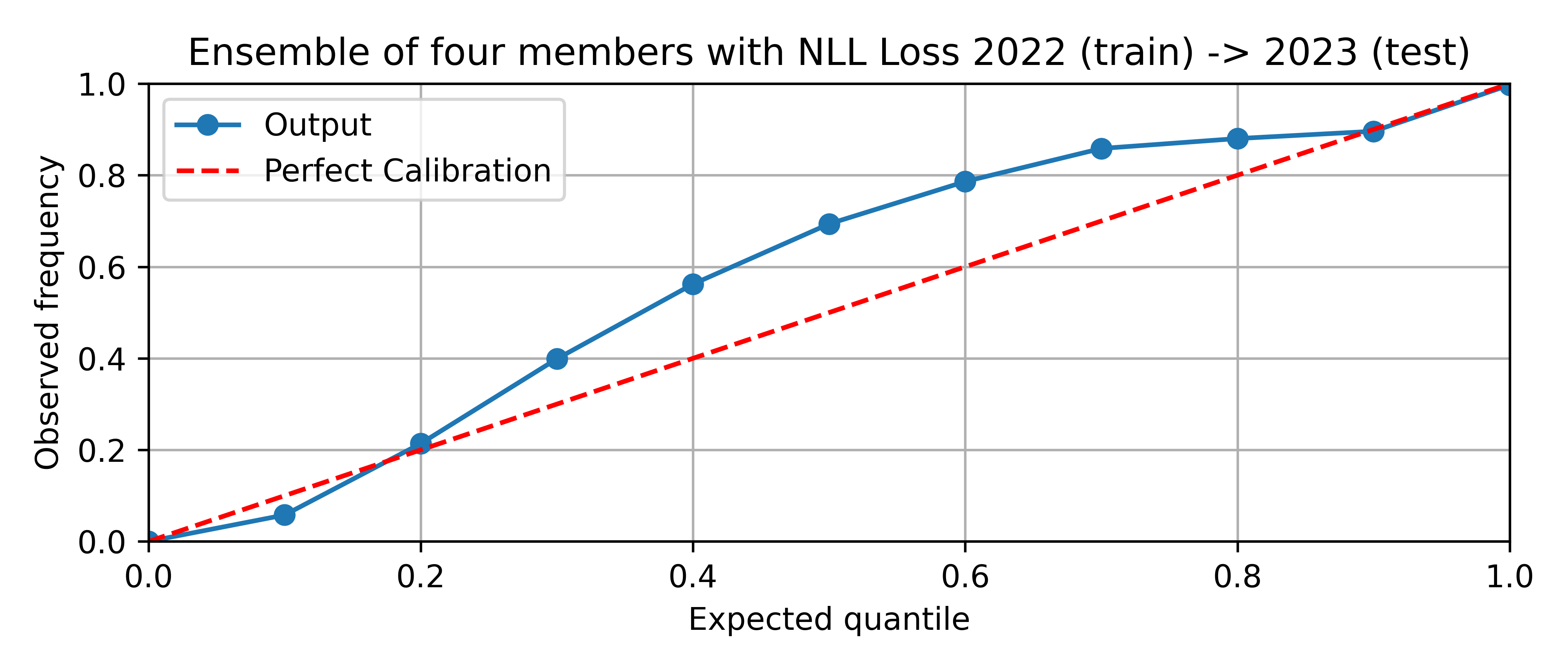}     
        \caption{}
        \label{fig:calib_on_2022}
    \end{subfigure}
    \hfill
    \begin{subfigure}{.48\linewidth}
        \centering
        \includegraphics[width=\linewidth]{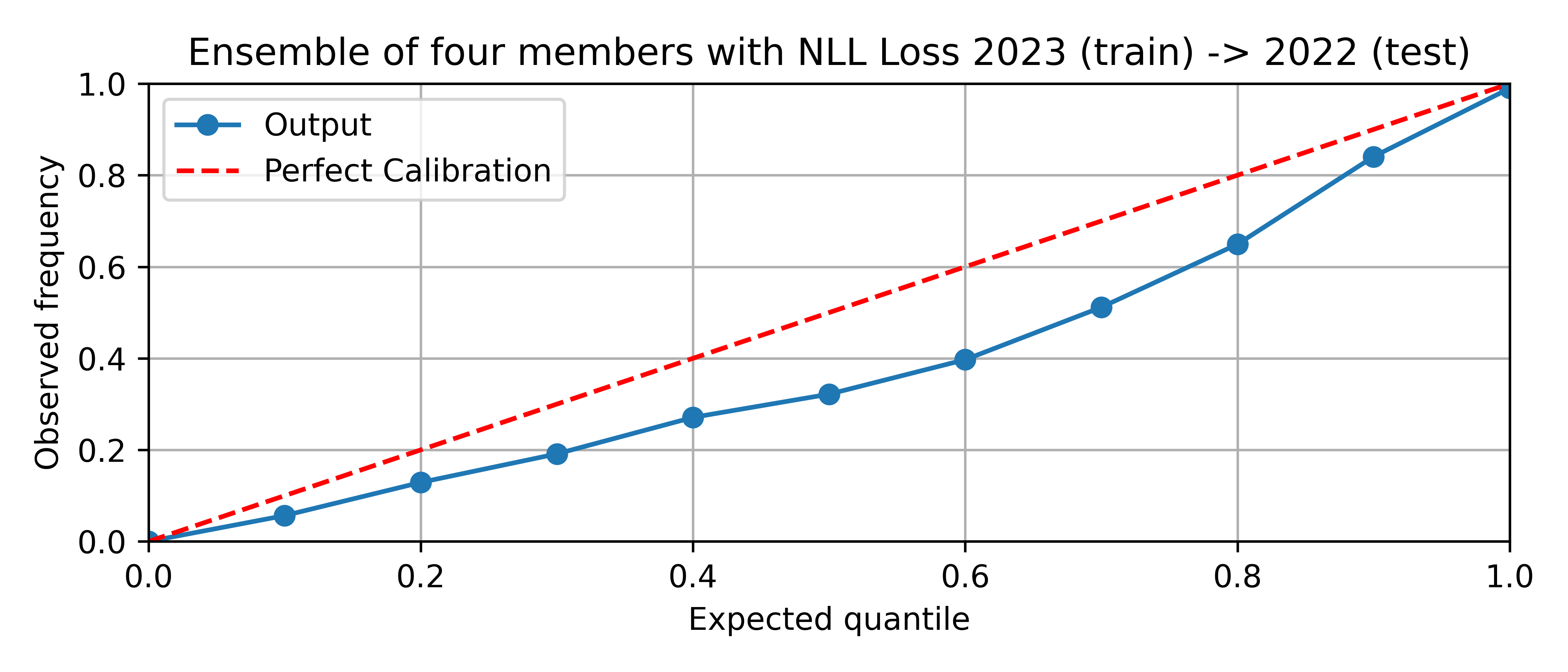}
        \caption{}
        \label{fig:calib_on_2023}
    \end{subfigure}
    \caption{Calibration assessment of an ensemble by using a reliability diagram. \textit{(a)}: Training on 2022 and testing on 2023 shows under-confidence, while \textit{(b)} demonstrates over-confidence.}
    \label{fig:calibration_results}
\end{figure*}

\subsection{Predictive uncertainty and calibration analysis}

Beyond point accuracy, a reliable forecasting system for greenhouse management must also produce well-calibrated uncertainty estimates, so that growers can distinguish confident predictions from uncertain ones and act accordingly. We measure calibration quantitatively using the Uncertainty Calibration Error (UCE)~\cite{Laves2020WellCalibratedRU} and NLL, reported in Table~\ref{tab:calibration}, and qualitatively through reliability diagrams shown in Figure~\ref{fig:calibration_results}.

\noindent UCE measures the discrepancy between a model's predicted uncertainty and its actual prediction error. To compute it, predictions are grouped into $B$ bins according to their predicted uncertainty level, and the average predicted uncertainty within each bin is compared against the average empirical error of that bin. The overall UCE is then the weighted mean of these per-bin discrepancies:

\begin{equation}
    \text{UCE} = \sum_{b=1}^{B} \frac{|B_b|}{N} \left| \overline{\sigma}^2_b - \overline{\text{err}}_b \right|
\end{equation}

\noindent where $|B_b|$ is the number of samples in bin $b$, $\overline{\sigma}^2_b$ is the mean predicted variance in that bin, and $\overline{\text{err}}_b$ is the mean squared error of the predictions in that bin. A UCE of zero indicates perfect calibration, meaning predicted uncertainty faithfully tracks actual error, while a value approaching one indicates severe miscalibration. The reliability diagram provides a visual counterpart to UCE, plotting $\overline{\sigma}^2_b$ against $\overline{\text{err}}_b$ for each bin; a perfectly calibrated model follows the diagonal line.

\begin{table}[t]
    \centering
    \begin{tabular}{l|cc}
         & UCE  & NLL  \\
         \midrule
         2022 $\rightarrow$ 2023 & \textbf{0.391} & \textbf{1.264} \\
         2023 $\rightarrow$ 2022 & 0.888 & 1.958 \\
         
    \end{tabular}
    \caption{Calibration metrics for the NLL-trained multi-modal forecasting ensemble. The lower the calibration metric the better.}
    \label{tab:calibration}
\end{table}

\paragraph{Cross-season generalization:} When training on 2022 and testing on 2023, the model achieves well-calibrated uncertainty estimates with low NLL and UCE, indicating that the predicted uncertainty faithfully reflects the actual prediction error as shown in Figure~\ref{fig:calibration_results} (a). In the ranges from 0.0 to 0.2 and 0.8 to 1.0 on the x-axis, the model is aligned with the diagonal line demonstrating approximately perfect calibration. However in the range of 0.2 to 0.8, the reliability diagram shows that the observed coverage consistently exceeds the expected coverage, the curve lies above the diagonal, indicating that the model is under-confident in this regime, producing prediction intervals that are wider than necessary. We observe the opposite when training on 2023 data and testing on 2022. The model is over-confident across the full range from 0.0 to 1.0 on the x-axis.

\subsection{Discussion}
Our results are consistent with the hypothesis that past plant states are predictive of future yield, and that modeling this temporal dependency explicitly, rather than treating observations independently, provides a substantial improvement over a persistence baseline. 
The contribution of visual information, extracted via self-supervised DinoV3 features, was modest: across both seasons the Multimodal model provided a consistent but small improvement over the Counting-only model, suggesting that visual features carry a complementary signal beyond fruit counts, though counts alone already capture most of the predictable variance.
The robustness of the leave-one-year-out cross-validation is particularly encouraging for applications. Despite the stochastic nature of individual plant development and the fluctuations introduced by human harvest intervention, the model effectively generalizes across seasons. This suggests that the integration of temporal deltas ($\Delta_t$) allows the network to successfully modulate biological transitions even when data acquisition is irregular.

Beyond point accuracy, the adoption of Gaussian NLL training within the ensemble framework enables the model to produce calibrated uncertainty estimates alongside its predictions. The calibration results in Table~\ref{tab:calibration} reveal an asymmetry between the two cross-season scenarios: training on 2022 and evaluating on 2023 yields lower UCE and NLL than the reverse, suggesting that the 2023 season provides a less representative prior for the distributional characteristics of 2022. This is further corroborated by the reliability diagrams in Figure~\ref{fig:calibration_results}, where the model trained on 2023 shows under-confidence.

Nevertheless, the cross-season UCE values remain low enough to confirm that the ensemble's uncertainty estimates are practically useful, providing growers with a principled confidence signal to support harvest decision-making under seasonal variability.

\section*{Conclusions}
In this work, we addressed the challenge of yield forecasting for individual plants by leveraging multimodal time-series data from sweet pepper crops. Our findings validate that plant development is a temporally dependent process where combining past numerical states with visual cues provides a more comprehensive representation of growth than discrete counts alone. 
Experimental results across two growing seasons confirmed the robustness of our framework, showing that performance improves with increased historical context. 

\noindent Additionally, by incorporating Deep Ensembles and Gaussian NLL, we provided a mechanism for uncertainty quantification, which is essential for building trust in automated agricultural systems. While cross-season generalization revealed potential for variance inflation in specific scenarios, the overall low Uncertainty Calibration Error (UCE) indicates that these estimates remain practically useful for growers. 
Ultimately, this research provides a scalable foundation for data-driven yield monitoring and logistics planning in controlled-environment horticulture.

\section*{CRediT authorship contribution statement}
\textbf{Enrico Pallotta}: Conceptualization, Data curation, Formal analysis, Investigation, Methodology, Software, Validation, Visualization, Writing - original draft, Writing - review \& editing.
\textbf{Mohamed Farag}: Formal analysis, Methodology, Software, Validation, Visualization, Writing - original draft, Writing - review \& editing.
\textbf{Esra Guclu}: Data curation, Resources, Visualization, Writing - review \& editing.
\textbf{Chris McCool}: Funding acquisition, Supervision, Writing - review \& editing.
\textbf{Ribana Roscher}: Funding acquisition, Supervision, Writing - review \& editing.
\textbf{Juergen Gall}: Funding acquisition, Project administration, Supervision, Writing - review \& editing.

\section*{Declaration of competing interest}
The authors declare that they have no known competing financial interests or personal relationships that could have appeared to influence the work reported in this paper.

\section*{Acknowledgments}
This work was supported by the Deutsche Forschungsgemeinschaft (DFG, German Research Foundation) GA 1927/9-1, RO 4839/6-1, and MC 831/2-1 (KI-FOR 5351 AID4Crops).
The authors gratefully acknowledge the access to the Marvin cluster at the University of Bonn, which enabled them to perform all experiments.

\bibliographystyle{cas-model2-names}
% \bibliographystyle{unsrt}

\end{document}